# MAPROC at AHaSIS Shared Task: Few-Shot and Sentence Transformer for Sentiment Analysis of Arabic Hotel Reviews

**Randa Zarnoufi**
Mohammed V University in Rabat
randa_zarnoufi@um5.ac.ma

**Abstract**

Sentiment analysis of Arabic dialects presents significant challenges due to linguistic diversity and the scarcity of annotated data. This paper describes our approach to the AHaSIS shared task, which focuses on sentiment analysis on Arabic dialects in the hospitality domain. The dataset comprises hotel reviews written in Moroccan and Saudi dialects, and the objective is to classify the reviewers' sentiment as positive, negative, or neutral. We employed the SetFit (Sentence Transformer Fine-tuning) framework, a data-efficient few-shot learning technique. On the official evaluation set, our system achieved an F1 of 73%, ranking 12th among 26 participants. This work highlights the potential of few-shot learning to address data scarcity in processing nuanced dialectal Arabic text within specialized domains like hotel reviews.

## 1 Introduction

Sentiment Analysis (SA), a key area within Natural Language Processing (NLP), focuses on identifying and extracting subjective information, such as opinions, emotions, and attitudes, from textual data. The proliferation of user-generated content on social media, review platforms, and forums has underscored the importance of robust sentiment analysis systems across various domains. One such domain is the hospitality industry, where customer reviews significantly influence consumer decisions and provide valuable feedback for service improvement. However, performing sentiment analysis on Arabic text, particularly regional dialects, presents unique and substantial challenges.

Arabic is a morphologically rich language characterized by a complex linguistic landscape. It encompasses Modern Standard Arabic (MSA), used in formal communication and written media, and a diverse array of regional dialects spoken in daily interactions. These dialects often differ significantly from MSA and from each other in terms of syntax, lexicon, phonology, and semantics. This linguistic diversity poses a considerable hurdle for NLP tasks, as models trained on MSA or one specific dialect may not generalize well to others. In the context of sentiment analysis, this variability is further compounded because emotional expressions, idiomatic phrases, and cultural nuances can vary widely across different Arabic-speaking regions, making consistent sentiment detection a difficult endeavor.

The AHaSIS (Arabic Hotel Reviews Analysis for Sentiment Identification in Dialects) shared task (Alharbi, Chafik, et al., 2025) aims to address these challenges by focusing on advancing sentiment analysis techniques specifically for Arabic dialects within the hotel review domain. The primary objective of this task is to classify the overall sentiment expressed in hotel reviews, written in various Arabic dialects, into three categories: positive, neutral, or negative. This task encourages participants to develop models capable of handling the intricacies of dialectal Arabic and accurately discerning sentiment despite the linguistic variations.

This dataset aims to serve as a multidialectal benchmark for sentiment analysis in the hospitality sector, helping to address the shortage of sentiment analysis resources in dialectal Arabic. While previous efforts have contributed valuable benchmarks for SA Arabic benchmarks; such as LABR (Aly & Atiya, 2013) for book reviews, ASTD (Nabil et al., 2015) for Arabic tweets SA, ArSentD-LEV (Baly et al., 2019) for the Levantine dialect on Twitter and Arsen-20 (Fang & Xu, 2024) that focuses on the theme of COVID-19; we still face a pressing need for more diverse and domain-specific datasets.

This paper describes our approach to the AHaSIS shared task. We present a solution that leverages the Huggingface SetFit[1] (Sentence Transformer Fine-tuning) framework, an efficient method for few-shot learning without requiring extensive prompt engineering and faster to train and run inference with, compared to other few-shot learning methods.

The core contribution of our work lies in demonstrating the effectiveness of SetFit for rapidly developing a competitive sentiment analysis model for low-resource dialectal Arabic, achieving promising results with limited training data.

We detail our data preprocessing steps, model configuration, training strategy, and the experimental results obtained on the shared task's dataset in the following sections.

## 2 Related Work

Sentiment analysis in Arabic has been an active area of research, driven by the increasing volume of Arabic content online and the need for tools to understand public opinion and customer feedback. Early approaches often relied on lexicon-based methods, which utilize dictionaries of words tagged with sentiment polarities such as ArsenL (Arabic Sentiment Lexicon) (Badaro et al., 2014) and Arabic Senti-Lexicon (Al-Moslmi et al., 2018). While straightforward, these methods struggle with the nuances of language, context dependent sentiment, and the morphological richness of Arabic (Mulki et al., 2017). Machine learning techniques, including traditional methods like Support Vector Machines (SVM) (Duwairi et al., 2015), Naive Bayes (Al-Horaibi & Khan, 2016), and Logistic Regression (Alshammari & AlMansour, 2020)have been widely applied to Arabic sentiment analysis. These approaches typically require significant feature engineering, such as n-grams, TF-IDF (Al-Osaimi & Badruddin, 2014; Salameh et al., 2015) and word embeddings. The development of Arabic-specific word embeddings, like AraVec (Soliman et al., 2017), has improved the performance of these models by capturing semantic relationships between words (Ashi et al., 2019). After the introduction of deep learning models, such as CNN and those based on RNNs like LSTMs and GRUs have advanced the SA performance (Al-Sallab et al., 2017; Al-Smadi et al., 2017; Dahou et al., 2016)

More recently, Transformer-based architectures such as AraBERT (Antoun et al., 2020), CAMeLBERT (Inoue et al., 2021), and ARBERT/MARBERT (Abdul-Mageed et al., 2021), which are specifically trained on large Arabic and dialectal corpora, have achieved state-of-the-art results in various NLP tasks, including SA. Pre-training language models for Arabic and fine-tuning it for SA, significantly improves the performance (Eljundi et al., 2019). These models can learn contextual representations of text, reducing the need for manual feature engineering.

Despite these advancements, SA for Arabic dialects remains a significant challenge (Shi & Agrawal, 2025). Most existing resources and models are primarily focused on Modern Standard Arabic (MSA) or a limited set of well-resourced dialects (Mashaabi et al., 2024). The linguistic diversity across dialects, including variations in vocabulary, grammar, and idiomatic expressions, makes it difficult to develop universally applicable sentiment analysis tools. Furthermore, the scarcity of large, annotated datasets for many Arabic dialects hinders the development and evaluation of robust dialectal sentiment analysis models.

Few-shot learning techniques have emerged as a promising direction for addressing data scarcity in NLP. Methods like SetFit that stands for Sentence Transformer Finetuning (Tunstall et al., 2022) which is employed in our work, aim to achieve strong performance with minimal labeled training examples. SetFit leverages sentence transformers to generate high-quality embeddings and then fine-tunes a classification head, offering an efficient alternative to training large models from scratch or relying on complex prompting strategies. SetFit uses supervised contrastive learning that has proved its efficiency in previous works (Khosla et al., 2020).

Our work builds upon these advancements by applying SetFit to the specific challenge of sentiment analysis in diverse Arabic dialects within the hotel review domain, contributing to the growing body of research on low-resource NLP and dialectal Arabic processing.

---

[1] https://huggingface.co/docs/setfit/en/index

| Dialect | Sentence | Polarity |
|---|---|---|
| Darija | عطاوني حتى كيكة فعيد ميلادي منين كنت جالس تماك. يستاهل كل فلس حطيتيه<br>They even gave me a cake on my birthday while I was sitting there. It deserves every penny you spent | positive |
| Darija | الأتمنة دلوطيل طالعة بزاف، وما كاين حتى مقابل فالسيرفيس ولا باش يرضيو الكليان<br>The hotel prices are way too high, and there's nothing in the service to match them or to please the customers | negative |
| Darija | فطور ما بيهش، يمكن يكون احسن<br>The breakfast wasn't bad, but it could be better | neutral |
| Saudi | غرف الفندق اطلالتها ساحرة ، نظافة الغرف مع دورات المياه وكل شي في الفندق حلو مره<br>The hotel rooms have a stunning view, the rooms and bathrooms are clean, and everything about the hotel is just great | positive |
| Saudi | غالي مره ملعب اطفال صغير فريق ترفيه خايس<br>Very expensive, the kids' play area is small, and the entertainment team is terrible | negative |
| Saudi | فندق لازمه اهتمام ممكن يكون جيد بس يحتاج تعديلات كثيرة<br>The hotel needs attention, it could be good, but it requires a lot of improvements. | neutral |

Table 1: Data Examples from the Training Set.

| Dataset | Size | Sentiment Distribution |
|---|---|---|
| Training | 860 samples: 430 samples for each dialect (Darija/Saudi) | Positive: 308 = 35.81%<br>Negative: 336 = 39.06%<br>Neutral: 216 = 25.11% |
| Evaluation | 216 samples: 108 for each dialect | — |

Table 2: Overall Data Distribution

| Dialect | Sentiment Distribution |
|---|---|
| Darija/Saudi | Positive: 154 = 35.81%<br>Negative: 168 = 39.06%<br>Neutral: 108 = 25.11% |

Table 3: Sentiment Distribution over Dialects in the Training Data.

## 3 Data

The primary dataset for this work was provided as part of the AHaSIS shared task. The dataset consists of hotel reviews written in various Arabic dialects, namely Moroccan dialect known as Darija and Saudi dialect. Some reviews are shown in Table 1 for both dialects.

The training set is composed of 860 samples, and the evaluation set contains 216 samples. The training set contains four columns for 'ID' of review, 'Sentiment' (the sentiment label: positive, neutral, or negative), 'Text' (the review text), and 'Dialect' (Darija or Saudi). The overall data distribution and sentiment class distribution over dialects are described in Table 2 and Table 3 respectively. We can observe that data distribution and the sentiment distribution over dialects are perfectly balanced 50% for each one. Although the sentiment classes (positive, negative, neutral) are not fully balanced, the differences in their distribution are relatively small and unlikely to introduce significant bias into the overall classification.

## 4 System Description

Our approach to the AHaSIS shared task on sentiment analysis for Arabic dialects in hotel reviews is centered around the SetFit (Sentence Transformer Fine-tuning) framework. SetFit is designed for efficient few-shot learning, enabling the development of robust text classification models with a small number of labeled examples, which are particularly beneficial for low-resource languages or specialized domains like dialectal Arabic.

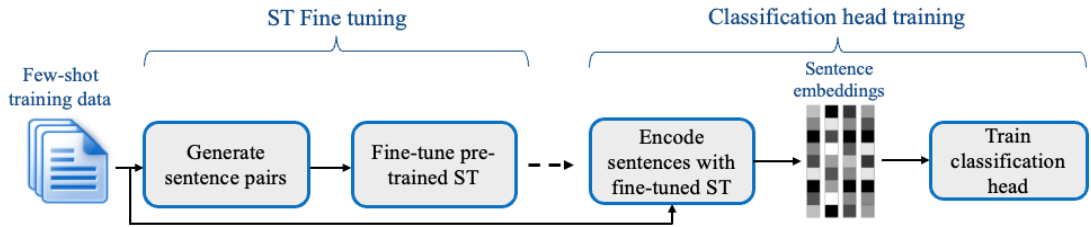

Figure 1: The SetFit framework consists of two processing stages: the first involves fine-tuning a sentence transformer, and the second trains a classification head on the resulting embeddings.

### 4.1 SetFit Framework Overview

The SetFit framework is composed of two main parts: the body which is a sentence transformer model, and a classification head as it is shown in Figure 1[2]. It operates in two stages. First, it fine-tunes the pre-trained sentence transformer model using a contrastive learning objective on the provided training data. This stage aims to generate rich sentence embeddings directly from text examples that are well-suited for the specific task and dataset. In this process, positive pairs (similar sentences) and negative pairs (dissimilar sentences) are created from the training examples to teach the model to produce similar embeddings for sentences with the same label and dissimilar embeddings for sentences with different labels. The second stage involves training a classification head (typically a logistic regression model or a simple neural network) on top of the frozen embeddings generated by the fine-tuned Sentence Transformer. This two-stage process allows SetFit to achieve strong performance without requiring extensive labeled data or complex prompt engineering, which are often associated with large language models.

### 4.2 Sentence Transformer Model

For our implementation, we utilized a pre-trained Arabic sentence transformer model Arabic-SBERT-100K[3]. This model is a variant of SBERT (Sentence BERT), that it is based on AraBERT base model and has been finetuned on a large corpus of Arabic text (100K sentences), making it suitable for generating meaningful sentence representations for Arabic. The choice of this model was based on its reported performance in comparison with other models and its availability on the Hugging Face Model Hub. For comparison purposes, we also used another sentence transformer model which is ArabicBERT_Finetuned-AR-500[4]

### 4.3 Baseline Model

We initiated our baseline by fine-tuning AraBERTv0.2 base model[5] on the full training dataset and then explored ways to enhance classification performance by varying the number of training samples.

### 4.4 Dataset

For our experiments, we used the provided train.csv file. While analyzing the data we have observed that it has been already pre-processed and doesn't need further pre-processing except removing punctuation and normalizing some Arabic letters like alif and hamza writing format ('ا', 'أ', 'إ').
We performed an 80/20 split on this dataset to create our internal training and testing partitions. For the few-shot learning aspect of SetFit, a smaller training subset was created by sampling a specific number of examples per class from the training set.

Our final reported model utilized 64 examples per sentiment class for fine-tuning the sentence transformer and training the classification head. This means the actual training data used by SetFit was relatively small.

---

[2] https://huggingface.co/blog/setfit
[3] https://huggingface.co/akhooli/Arabic-SBERT-100K
[4] https://huggingface.co/danfeg/ArabicBERT_Finetuned-AR-500
[5] https://huggingface.co/aubmindlab/bert-base-arabertv2

| Model | Samples Number | Epoch | F1 % | Duration in h:m:s |
|---|---|---|---|---|
| Baseline: finetuned AraBERT | Full training dataset | 3 | 75.07 | 00:17:10 |
| Arabic-SBERT-100K | 8 | 1 | 63.82 | 00:05:12 |
| | 8 | 3 | 67.48 | 00:15:14 |
| | 8 | 5 | 64.22 | 00:26:09 |
| | 16 | 3 | 69.19 | 00:30:54 |
| | 32 | 3 | 73.84 | 01:04:35 |
| | 64 | 3 | **78.87** | 02:16:12 |
| ArabicBERT_Finetuned-AR-500 | 8 | 3 | 52.83 | 00:43:32 |

Table 4: Results on the AHaSIS Training dataset.

| Model | Samples Number | Epoch | F1 |
|---|---|---|---|
| Arabic-SBERT-100K | 32 | 3 | 72% |
| | 64 | 3 | **73%** |

Table 5: Results on the AHaSIS Evaluation dataset.

## 4.5 Training Procedure

The training process followed the standard SetFit pipeline. We selected a subset of the training data for the few-shot learning setup. We experimented with different numbers of training examples per class.

The SetFitTrainer was configured with the following key parameters:

**Model**: The Arabic-SBERT-100K SetFit model.

**Training Dataset**: The few-shot training dataset containing *n* examples per class.

**Test Dataset**: The full test dataset derived from the initial data split.

**Number of Epochs**: The sentence transformer was fine-tuned for 3 epochs during the contrastive learning phase. This value was chosen based on preliminary experiments, which showed that 3 epochs provided a good balance between performance and training time for this specific model and dataset size.

**Batch Size**: A batch size of 16 was used during the fine-tuning of the sentence transformer.

**Number of Iterations**: 20 iterations were used for generating text pairs for contrastive learning.

**Loss Function**: CosineSimilarityLoss was employed for the contrastive fine-tuning stage, which is a standard choice for SetFit to encourage similar sentences to have embeddings with high cosine similarity.

After the Sentence Transformer was fine-tuned, the classification head was trained on the embeddings generated from the selected training set samples.

For the final reported model, we used 64 examples per class from the training set to fine-tune the sentence transformer and train the classification head. This corresponds to the shot train set created by sampling 64 instances for each label ('positive', 'negative', 'neutral').

## 5 Results and Discussion

This section details the results achieved by our SetFit-based sentiment analysis model for Arabic hotel reviews. We mentioned that for our experimental settings, all experiments were conducted using CPU (i7-11850H) with 8 cores and 32GB of RAM to prove that this approach is not computationally expensive. The timing of each experience is also reported in Table 4.

### 5.1 Results

The primary metric for the shared task is F1 score.

The key results relevant to our submitted approach are shown in Table 5 and are as follows:

Using Arabic-SBERT-100K with 64 samples per class and 3 epochs for fine-tuning we achieved a F1 score of 73% in the official evaluation set, significantly outperforming the shared task baseline of 56% (Alharbi, Ezzini, et al., 2025).

We report also the specific results achieved with different configurations based on our internal test set split in Table 4 and for each dialect and class in Table 6.

These results demonstrate a clear trend:
- Increasing the number of few-shot samples per class, up to 64 in these experiments, generally improves the model's

| Dialect | Class | F1 % |
|---|---|---|
| Darija | positive | 72.92 |
| | negative | **85.85** |
| | neutral | 61.30 |
| Saudi | positive | **87.52** |
| | negative | 85.55 |
| | neutral | 80.00 |

Table 6: Results per Dialect and Class on the AHaSIS Training dataset.

performance on the internal test set. The choice of 3 epochs for fine-tuning the sentence transformer also appears to be beneficial compared to a single epoch and five epochs when using a small number of samples.
- The results obtained using the SetFit framework with the Arabic-SBERT-100K sentence transformer model were better than those with ArabicBERT_Finetuned-AR-50.
- Our internal experiments, culminating in an F1 of 78.87% on our held-out test set when using 64 samples per class, suggest that SetFit can effectively leverage pre-trained Arabic language representations for this task with relatively minimal training data.

One of the key strengths of our approach is its efficiency. Traditional deep learning models often require substantial amounts of labeled data and extensive computational resources for training. SetFit, by fine-tuning a sentence transformer with a contrastive objective and then training a simple classification head, offers a more resource-friendly alternative.

The ability to achieve competitive performance with only 64 examples per class (a total of 192 training examples for three classes) underscores the data efficiency of this method.

We mentioned that the official leaderboard on the shared task website [6] and Codabench show various teams' F1 scores, providing a benchmark for performance.

### 5.2 Discussion

The progressive improvement in F1 as the number of few-shot samples per class increased (from 8 to 64 samples) aligns with expectations. More data, even in a few-shot context, generally allows the model to learn more robust representations and decision boundaries. The choice of 3 epochs for fine-tuning the sentence transformer also appeared to be beneficial, suggesting that even a brief period of contrastive learning can adapt the pre-trained embeddings effectively to the target task and domain.

Comparing our F1 score of 73% with the best F1 scores on the official AHaSIS leaderboard (where the top score is 81% F1), our model appears to be competitive. Several factors could influence the performance. The quality of the pre-trained sentence transformer is crucial. Its training on a substantial Arabic corpus likely provides a good foundation for understanding Arabic semantics and syntax. The diversity of dialects present in the AHaSIS dataset is a significant challenge. While SetFit helps, the model's ability to generalize across highly varied dialectal expressions might still be limited by the relatively small fine-tuning dataset. The hotel review domain also has its specific vocabulary and expressions of sentiment, which the model needs to learn.

As we mentioned, we had also evaluated the model on our internal test set for each dialect and class. The results in Table 6 show that the model performs best on the Saudi dialect, achieving over 80% F1 in all classes, including strong handling of neutral sentiment (80%). In contrast, it struggles more with Darija, especially in the neutral class (61.3%), suggesting difficulty in capturing subtler expressions in that dialect which may be underrepresented in the sentence transformer training data.

In our experiments we also tested an alternative model, ArabicBERT_Finetuned-AR-500, which achieved a lower F1 (52.83 % with 3 epochs). This highlights the importance of selecting an appropriate base sentence transformer model for the SetFit framework. Arabic-SBERT-100K model seems better suited for this specific task based on these preliminary results.

---

[6] https://Ahasis-42267.web.app/leaderboard

Finally, A detailed error analysis on the predictions would be valuable to understand the types of reviews or dialects where the model performs poorly, guiding further improvements.

# 6 Conclusion

In this paper, we presented our solution for the AHaSIS shared task, which focused on sentiment analysis of Arabic hotel reviews written in various dialects. Our approach utilized the SetFit framework, a few-shot learning technique. This methodology was chosen for its efficiency in scenarios with limited labeled data, a common challenge in processing diverse Arabic dialects.

Our system achieved an F1 of 73% in the official evaluation set. This performance suggests that few-shot learning with appropriate pre-trained models is a viable strategy for tackling sentiment analysis in complex linguistic landscapes like dialectal Arabic. The results also indicated a positive correlation between the number of few-shot samples and model performance, within the tested range.

Future work could explore several avenues for improvement. Experimenting with a wider range of pre-trained Arabic or multilingual sentence transformers, conducting a more extensive hyperparameter optimization for the SetFit trainer, and increasing the number of few-shot training examples could potentially enhance performance.

# References

Abdul-Mageed, M., Elmadany, A., Moatez, E., & Nagoudi, B. (2021). *ARBERT & MARBERT: Deep Bidirectional Transformers for Arabic*. https://github.com/attardi/wikiextractor.

Alharbi, M., Chafik, S., Ezzini, S., Mitkov, R., Ranasinghe, T., & Hettiarachchi, H. (2025). AHaSIS: Shared Task on Sentiment Analysis for Arabic Dialects. *Proceedings of the 15th International Conference on Recent Advances in Natural Language Processing (RANLP)*.

Alharbi, M., Ezzini, S., Hettiarachchi, H., Ranasinghe, T., & Mitkov, R. (2025). Evaluating Large Language Models on Arabic Dialect Sentiment Analysis. *Proceedings of the 15th International Conference on Recent Advances in Natural Language Processing (RANLP)*.

Al-Horaibi, L., & Khan, M. B. (2016). Sentiment analysis of Arabic tweets using text mining techniques. *International Workshop on Pattern Recognition*. https://api.semanticscholar.org/CorpusID:59094364

Al-Moslmi, T., Albared, M., Al-Shabi, A., Omar, N., & Abdullah, S. (2018). Arabic senti-lexicon: Constructing publicly available language resources for Arabic sentiment analysis. *Journal of Information Science*, *44*(3), 345–362. https://doi.org/10.1177/0165551516683908

Al-Osaimi, S., & Badruddin, K. M. (2014). Role of Emotion icons in Sentiment classification of Arabic Tweets. *Proceedings of the 6th International Conference on Management of Emergent Digital EcoSystems*, 167–171. https://doi.org/10.1145/2668260.2668281

Al-Sallab, A., Baly, R., Hajj, H., Shaban, K. B., El-Hajj, W., & Badaro, G. (2017). AROMA: A Recursive Deep Learning Model for Opinion Mining in Arabic as a Low Resource Language. *ACM Trans. Asian Low-Resour. Lang. Inf. Process.*, *16*(4). https://doi.org/10.1145/3086575

Alshammari, N. F., & AlMansour, A. A. (2020). *Aspect-based Sentiment Analysis for Arabic Content in Social Media*.

Al-Smadi, M., Qawasmeh, O., Al-Ayyoub, M., Jararweh, Y., & Gupta, B. B. (2017). Deep Recurrent neural network vs. support vector machine for aspect-based sentiment analysis of Arabic hotels' reviews. *J. Comput. Sci.*, *27*, 386–393. https://api.semanticscholar.org/CorpusID:51918752

Aly, M., & Atiya, A. (2013). LABR: A Large Scale Arabic Book Reviews Dataset. In H. Schuetze, P. Fung, & M. Poesio (Eds.), *Proceedings of the 51st Annual Meeting of the Association for Computational Linguistics (Volume 2: Short Papers)* (pp. 494–498). Association for Computational Linguistics. https://aclanthology.org/P13-2088/

Antoun, W., Baly, F., & Hajj, H. (2020). *AraBERT: Transformer-based Model for Arabic Language Understanding*. http://arxiv.org/abs/2003.00104

Ashi, M. M., Siddiqui, M. A., & Nadeem, F. (2019). *Pre-trained Word Embeddings for Arabic Aspect-Based Sentiment Analysis of Airline Tweets* (A. E. Hassanien, M. F. Tolba, K. Shaalan, & A. T. Azar, Eds.; Vol. 845). Springer International Publishing. https://doi.org/10.1007/978-3-319-99010-1

Badaro, G., Baly, R., Hajj, H., Habash, N., & El-Hajj, W. (2014). A Large Scale Arabic Sentiment Lexicon for Arabic Opinion Mining. In N. Habash & S. Vogel (Eds.), *Proceedings of the EMNLP 2014 Workshop on Arabic Natural Language Processing (ANLP)* (pp. 165–173). Association for Computational Linguistics. https://doi.org/10.3115/v1/W14-3623

Baly, R., Khaddaj, A., Hajj, H. M., El-Hajj, W., & Shaban, K. B. (2019). ArSentD-LEV: A Multi-Topic Corpus for Target-based Sentiment Analysis in Arabic Levantine Tweets. *CoRR*, *abs/1906.01830*. http://arxiv.org/abs/1906.01830

Dahou, A., Xiong, S., Zhou, J., Haddoud, M. H., & Duan, P. (2016). Word Embeddings and Convolutional Neural Network for Arabic Sentiment Classification. In Y. Matsumoto & R. Prasad (Eds.), *Proceedings of COLING 2016, the 26th International Conference on Computational Linguistics: Technical Papers* (pp. 2418–2427). The COLING 2016 Organizing Committee. https://aclanthology.org/C16-1228/

Duwairi, R., Ahmed, N. A., & Al-Rifai, S. Y. (2015). Detecting sentiment embedded in Arabic social media - A lexicon-based approach. *J. Intell. Fuzzy Syst.*, *29*, 107–117. https://api.semanticscholar.org/CorpusID:6145765

Eljundi, O., Antoun, W., Nour, ), Droubi, E., Hajj, H., El-Hajj, W., & Shaban, K. (2019). *hULMonA ( ): The Universal Language Model in Arabic*.

Fang, Y., & Xu, C. (2024). ArSen-20: A New Benchmark for Arabic Sentiment Detection. *5th Workshop on African Natural Language Processing*. https://openreview.net/forum?id=GgsRUF5kJt

Inoue, G., Alhafni, B., Baimukan, N., Bouamor, H., & Habash, N. (2021). *The Interplay of Variant, Size, and Task Type in Arabic Pre-trained Language Models*. http://arxiv.org/abs/2103.06678

Khosla, P., Teterwak, P., Wang, C., Sarna, A., Tian, Y., Isola, P., Maschinot, A., Liu, C., & Krishnan, D. (2020). *Supervised Contrastive Learning*. http://arxiv.org/abs/2004.11362

Mashaabi, M. O., Al-Khalifa, S. Z., & Al-Khalifa, H. S. (2024). *A Survey of Large Language Models for Arabic Language and its Dialects*. https://mistral.ai/news/mistral-saba

Mulki, H., Haddad, * -Hatem, & Babaoˇ, I. (2017). *Modern Trends in Arabic Sentiment Analysis: A Survey* (Vol. 58).

Nabil, M., Aly, M., & Atiya, A. (2015). ASTD: Arabic Sentiment Tweets Dataset. In L. Màrquez, C. Callison-Burch, & J. Su (Eds.), *Proceedings of the 2015 Conference on Empirical Methods in Natural Language Processing* (pp. 2515–2519). Association for Computational Linguistics. https://doi.org/10.18653/v1/D15-1299

Salameh, M., Mohammad, S. M., & Kiritchenko, S. (2015). *Sentiment after Translation: A Case-Study on Arabic Social Media Posts*. http://www.purl.com/net/ArabicSentiment

Shi, Z., & Agrawal, R. (2025). *A comprehensive survey of contemporary Arabic sentiment analysis: Methods, Challenges, and Future Directions*.

Soliman, A. B., Eissa, K., & El-Beltagy, S. R. (2017). AraVec: A set of Arabic Word Embedding Models for use in Arabic NLP. *Procedia Computer Science*, *117*, 256–265. https://doi.org/10.1016/j.procs.2017.10.117

Tunstall, L., Reimers, N., Jo, U. E. S., Bates, L., Korat, D., Wasserblat, M., & Pereg, O. (2022). *Efficient Few-Shot Learning Without Prompts*. http://arxiv.org/abs/2209.11055